\pdfoutput=1
\documentclass[pmlr]{jmlr}
\usepackage{amsmath,amssymb,graphicx,url}
\usepackage{lmodern}
\usepackage{textcomp}
\renewcommand{\copyright}{\textcopyright}
\usepackage{booktabs}
\hypersetup{
  pdftitle={Conformal Expert Pools: Training-Free Probabilistic Time-Series Forecasting Beyond Seasonality},
  pdfauthor={Valery Manokhin},
  pdfsubject={Conformal prediction; probabilistic time-series forecasting},
  pdfkeywords={conformal prediction, probabilistic forecasting, time series, CRPS, Winkler score, MSIS, calibration, training-free baselines, forecast combination}
}

\jmlrvolume{}
\jmlryear{2026}
\jmlrworkshop{Preprint}
\jmlrproceedings{Preprint}{November 2026}

\title[Conformal Expert Pools]{%
Conformal Expert Pools: Training-Free Probabilistic\\
Time-Series Forecasting Beyond Seasonality}

\author{\Name{Valery Manokhin}\Email{Valery.Manokhin.2015@live.rhul.ac.uk}\\
  \addr{Independent researcher}}

\begin{document}
\maketitle

\begin{abstract}
Conformal Seasonal Pools (CSP; \citealp{manokhin2026csp}) showed that a forecaster with no
training at all, built only from past values at the same point in the season and from past
forecast errors, significantly outperformed DeepNPTS \citep{rangapuram2023deepnpts}, Amazon's
learned non-parametric forecaster. CSP was the more accurate of the two on $66\%$ of $232$
datasets, its $95\%$ prediction intervals contained the outcome $90.1\%$ of the time against
$57.8\%$ for DeepNPTS, and DeepNPTS was $819$ times slower. CSP was always meant to be the
simplest possible probabilistic forecasting baseline, a few lines of numpy code, and was never
designed as a full-fledged forecasting model. It forecasts the level of a series in one way only,
with the seasonal-naive forecast, and so leaves room for improvement on trending series, short
histories, changing volatility and series with many zeros.

We propose Conformal Expert Pools (CEP), a training-free successor. CEP replaces the single
seasonal-naive forecast by a pool of simple forecasters, builds its predictive distribution from
their past forecast errors, and adapts both the pool and the errors to the series being forecast.
No model is trained and no parameter is learned across series.

We evaluate CEP on the CSP benchmark of $232$ datasets and $24{,}502$ forecast windows, using
$95\%$ prediction intervals. We report two main probabilistic scores, for both of which lower is
better: CRPS \citep{gneiting2007strictly}, which measures the accuracy of the whole predictive
distribution, and the scaled interval score (MSIS; \citealp{makridakis2020m4}), which penalises
intervals that are wide or that miss the outcome. Compared with CSP, CEP is better on $78\%$ of
datasets by CRPS, on $84\%$ by MSIS and on $83\%$ by the unscaled Winkler score
\citep{winkler1972decision}, and on $78\%$ by the mean quantile loss used in the GluonTS and
DeepNPTS evaluations \citep{alexandrov2020gluonts,rangapuram2023deepnpts}. On a typical dataset its CRPS is $14\%$ lower and its
MSIS $23\%$ lower (both $p<10^{-22}$, paired Wilcoxon test; \citealp{wilcoxon1945individual}), and
interval coverage rises from $90.1\%$ to $92.2\%$. CEP also has a lower CRPS than DeepNPTS on
$88\%$ of datasets, than NPTS on $89\%$ and than SeasonalNPTS on $83\%$
\citep{alexandrov2020gluonts}. The improvement over CSP persists on the datasets from
families outside the development set (better on $76\%$ by CRPS, $83\%$ by MSIS and $81\%$ by Winkler score) and on $36$
additional files from fev-bench \citep{shchur2025fevbench} outside the benchmark ($78\%$, $81\%$
and $81\%$). Point forecasts improve as well: median MASE \citep{hyndman2006mase} falls from $1.10$ to
$1.01$ and median RMSSE \citep{makridakis2022m5u} from $0.94$ to $0.87$.

Finally, we compare CEP with Laplace \citep{cotton2026skaters}, an online ensemble forecaster that
borrowed from CSP: by its author's account \citep{cotton2026csppage}, after losing to CSP on
hourly data it adopted a ``seasonal anchor'', which blends a recency-weighted average of past
values at the same point in the season, the idea behind CSP's seasonal pool, with the
seasonal-naive value on which CSP centres its residuals. Its author states that ``Laplace beats (almost) everything'' \citep{cotton2026skaters} and
reports that it ``beats'' CSP on most series in three of four test collections
\citep{cotton2026laplacevscsp}. Those claims rest on a narrow test: one-step-ahead
forecasts only, on a limited and unrepresentative set of series chosen by Laplace's author (economic
series from one source and one hourly collection), with no interval widths or interval scores
reported. We tested them on the CSP benchmark, on $2{,}534$ multi-step windows from $228$ datasets
with the same history given to every method, and they do not hold on large-scale studies such as
this one. The first claim is false: plain CSP outperforms Laplace on $60\%$ of datasets by CRPS and on $64\%$ by interval
score, and its point forecasts are more accurate on $56\%$ by MASE. Scored
alongside all the methods of the benchmark on these datasets, Laplace ranks sixth of seven on both
CRPS and interval score: below SeasonalNPTS on both, below DeepNPTS on CRPS, and with by far the
widest intervals of all seven ($69\%$ wider than CEP's). Laplace's author did not report
interval widths, although it is generally accepted that a probabilistic forecast is judged on
sharpness (interval width) together with calibration (coverage)
\citep{gneiting2007calibration,gneiting2014probabilistic}; Laplace's intervals are $44\%$ wider
than CSP's.
Its weakness is at multi-step horizons:
from $13$ to $96$ steps CSP has the lower CRPS on $65\%$ of datasets. Its point forecast is
biased low (bias $-0.24$ in scaled units, against $-0.02$ for CEP and CSP) and on $54\%$ of datasets
is less accurate than the seasonal-naive forecast. Absence of bias is a key requirement of any
forecaster: a forecast that is systematically too low means stock-outs in demand planning and
under-procurement for an energy supplier bidding into the grid. Laplace fails this requirement. It also
produces pathological forecasts: on electricity load it abandons the daily cycle and forecasts
negative demand within a day, in $7.6\%$ of forecasts of non-negative series its median goes below
zero, and on some series its forecast distribution becomes numerically unstable, with intervals
fifteen orders of magnitude wider than the data and CRPS more than ten times CSP's on $9$ datasets. From a cold start Laplace takes about $7{,}700$ times longer than CSP, from $2{,}500$ times at
horizons up to $12$ to $24{,}000$ times beyond $96$, where a single forecast takes $107$ seconds, and
the longest horizons cannot be run in practice. CEP in turn is better than Laplace on
$82\%$ of datasets by CRPS, on $80\%$ by MSIS and on $82\%$ by Winkler score, its intervals are about $40\%$ narrower, and it
runs $271$ times faster.
\end{abstract}
\begin{keywords}
conformal prediction; probabilistic forecasting; time series; training-free baselines; forecast
combination; calibration; forecast accuracy; forecast evaluation; prediction intervals; CRPS;
interval score; Winkler score; MSIS; quantile loss; MASE; RMSSE; coverage
\end{keywords}

\section{Introduction}

A probabilistic forecast gives a range of plausible future values, not a single number. The usual
way to produce one is to train a model. \citet{manokhin2026csp} showed that training is not
always needed. Conformal Seasonal Pools (CSP) produces future scenarios in two simple ways: it
draws past values from the same point in the seasonal cycle (the same hour of the day, say), and
it adds past errors of the seasonal-naive forecast (``the same as one season ago'') to that
forecast. Prediction intervals are read directly from the scenarios. With no training and no
learned parameters, CSP was compared with DeepNPTS \citep{rangapuram2023deepnpts}, a learned
non-parametric forecaster from Amazon, on $232$ datasets. That comparison found three things.
\begin{itemize}
\item \emph{Accuracy.} CSP had the lower CRPS on $66\%$ of the datasets.
\item \emph{Coverage.} DeepNPTS's $95\%$ intervals contained the outcome only $57.8\%$ of the time,
against $90.1\%$ for CSP, and on many forecasts its coverage collapsed altogether: in $20.4\%$ of
forecast windows the DeepNPTS interval contained the outcome at none of the forecast steps, against
$1.5\%$ for CSP. On the six datasets on which DeepNPTS was originally introduced, the worst tenth
of its windows had coverage of $0.04$ or less.
\item \emph{Cost.} DeepNPTS was $819$ times slower: $5.52$ seconds per forecast against $6.75$
milliseconds, and $37.6$ hours of processor time to train $1{,}205$ networks, where CSP trains
nothing.
\end{itemize}
The paper argued that CSP should be a mandatory baseline.

CSP was designed from the start to be the simplest possible probabilistic forecasting baseline,
one that can be implemented in a few lines of numpy code. It was never intended as a full-fledged
forecasting model, and that simplicity sets its scope. Its only forecast of
where the series is heading is the seasonal-naive forecast. When a series trends, when the history is too short to contain
many values from the same point in the cycle, when volatility changes, or when most observations
are zero, errors of the seasonal-naive forecast are a less reliable guide to the future. In the
CSP benchmark this shows up as intervals that contain the outcome only $56\%$ of the time on histories
of $30$ points or fewer, and $82\%$ of the time at horizons up to $12$ steps, against a target of
$95\%$. Both groups are dominated by yearly, quarterly and monthly series.

This paper asks how far the training-free idea can be pushed without giving up what made CSP
useful: no fitting across series, a few lines of dependencies, and cost low enough to run as a
default. Our answer, Conformal Expert Pools (CEP), replaces the single seasonal-naive forecast by a
pool of simple forecasters and builds the predictive distribution from their past errors, adapted
to the series being forecast.

\paragraph{Contributions.}
\begin{enumerate}
\item A training-free probabilistic forecaster, CEP, that extends CSP beyond the
seasonal-naive forecast (Section~\ref{sec:methods}).
\item An evaluation on $232$ datasets with dataset-level inference, a held-out split by dataset
family, and a further test set of $36$ files from outside the corpus
(Sections~\ref{sec:design}--\ref{sec:results}).
\item A like-for-like comparison with the Laplace forecaster of \citet{cotton2026skaters} on
$2{,}534$ windows from $228$ datasets, which shows that published claims of Laplace's superiority over CSP do not hold on large-scale
studies such as the CSP benchmark (CSP wins on $60\%$ of datasets by CRPS and $64\%$ by interval score, with Laplace about
$7{,}700$ times slower), and in which CEP clearly outperforms Laplace: it has the lower
CRPS on $82\%$ of datasets, the lower interval score on $80\%$ (MSIS) and $82\%$ (Winkler) and the more accurate point forecast
(MASE) on $71\%$, while Laplace is more than two orders of magnitude slower ($271$ times;
Sections~\ref{sec:laplace}--\ref{sec:runtime}). The comparison also shows that Laplace exhibits
pathological behaviour: it forecasts negative values for non-negative series, loses the seasonal
cycle at multi-step horizons and becomes numerically unstable on some series. Its cost rules out
standard large-scale evaluations: at its measured speed Laplace would need about $52$
processor-hours to forecast the $100{,}000$ series of the M4 competition and about $105$
processor-hours for the $42{,}840$ series of the M5 competition, against $66$ and $130$ seconds
for CSP.
\end{enumerate}

\paragraph{How to read the results.} We score forecasts with CRPS and with the scaled interval
score (MSIS); for both, lower is better, and Section~\ref{sec:design} defines them. A method
``wins'' a dataset when its average score on that dataset is the lower one. Coverage is the share
of outcomes that fall inside the $95\%$ prediction interval, so the target is $0.95$.

We use ``conformal'' in the sense of the CSP paper: predictive distributions are built from
empirical pools of past non-conformity scores (signed forecast residuals). As in the CSP paper \citep{manokhin2026csp}, we make no formal coverage claim for CEP. Split-conformal constructions have been shown to remain approximately valid
under temporal dependence \citep{oliveira2024splitconformal,barber2025splitconformal}, and we
establish coverage empirically, in large-scale tests: at a nominal level of $95\%$ CEP's intervals
contained the outcome $92.2\%$ of the time over $24{,}502$ forecast windows from $232$ datasets.

\section{Related Work}

\paragraph{Training-free samplers.} NPTS and its seasonal variant sample from past values with a
recency or phase kernel \citep{alexandrov2020gluonts}; DeepNPTS learns the kernel
\citep{rangapuram2023deepnpts}. CSP \citep{manokhin2026csp} adds signed residual draws around the
seasonal-naive forecast. CEP belongs to this family of training-free samplers.

\paragraph{Conformal prediction for time series.} Split conformal prediction
\citep{vovk2005algorithmic,papadopoulos2007icp,lei2018distribution} calibrates on held-out
residuals; \citet{barber2025splitconformal} and \citet{oliveira2024splitconformal} show that its
coverage loss under temporal dependence is bounded for mixing processes, which supports
calibrating on residuals from the same series. Weighted variants down-weight old scores
\citep{barber2023nexcp}. Online methods adjust a threshold after each outcome: adaptive conformal
inference \citep{gibbs2021adaptive}, its aggregated form \citep{zaffran2022agaci}, conformal PID
control \citep{angelopoulos2023pid} and decaying step sizes \citep{angelopoulos2024decaying}.
Multi-step versions must handle delayed feedback and dependence across horizons
\citep{stankeviciute2021ctsf,szabadvary2024multistep,wang2024acmcp}. Ensemble approaches aggregate
conformal sets or calibrators with online weights \citep{xu2021enbpi,gasparin2024coma,liu2026pidexpert}.
CEP draws on this literature. It
differs in being a batch procedure that returns a full sample-based predictive distribution rather
than an interval wrapped around an external forecaster.

\section{Method}
\label{sec:methods}

This section describes CEP at the level of its design principles.

\paragraph{Input and output.} CEP takes the history of a single series, its seasonal period, a
forecast horizon and a nominal level. It returns a set of future scenarios (samples) and a
prediction interval for every step of the horizon. It uses no covariates and no other series.

\paragraph{From one forecaster to a pool.} CSP places past errors around one forecast of the
level of the series, the seasonal-naive forecast. CEP replaces that single forecast by a pool of
simple, classical point forecasters (``experts''). All of them have fixed settings. Nothing is fitted to the series and nothing is
learned across series.

\paragraph{Empirical error pools.} As in CSP, uncertainty is not modelled parametrically. The
predictive distribution is built from the errors the experts actually made on the history of the
same series, so the forecast inherits the empirical shape of those errors.

\paragraph{Adaptation.} CEP adapts to the series being forecast, using the history of that series
alone.

\paragraph{Training-free.} CEP has no training stage, no learned parameters and no
hyper-parameter search. The median time per
forecast on the corpus is $39$ ms on one core (Section~\ref{sec:runtime}).

\paragraph{What Laplace took from CSP.} CEP shares no code with Laplace.
Laplace took its seasonal anchor from its comparison with CSP. In July 2026 the Laplace of the time lost to CSP's default configuration on
$91\%$ of the M4-Hourly series by CRPS \citep{cotton2026laplacevscsp}. Its author traced the gap to
the seasonal location and added a \texttt{seasonal\_anchor} component to Laplace's default
candidates: an equal blend of a recency-weighted mean of past values at the same seasonal phase and
the value one season ago. The first is the quantity CSP's seasonal pool samples from; the second is
the seasonal-naive forecast around which CSP places its residuals. The change improved Laplace's
median CRPS on M4-Hourly by $8.3\%$, and the author's page on CSP states that Laplace ``adopted a
seasonal anchor as a result'' of that comparison \citep{cotton2026csppage}. CSP is not credited
correctly for this: neither the package's README nor the documentation of the component in the code
names CSP as its source.

\section{Experimental Design}
\label{sec:design}

\paragraph{Benchmark and protocol.} We reuse the CSP benchmark (the ``corpus'') without change
\citep{manokhin2026csp}: $232$ datasets and $24{,}502$ forecast windows. A \emph{window} is one
forecast: a series is cut at an origin, the method sees everything before the cut, and its
forecast is scored against the next $H$ observations. Every method produces $N=200$ samples and a
$95\%$ prediction interval ($\alpha=0.05$), with fixed random seeds. Results for CSP, CSP-Nixtla
\citep{nixtla2026statsforecast}, DeepNPTS \citep{rangapuram2023deepnpts}, NPTS, SeasonalNPTS
\citep{alexandrov2020gluonts} are the stored results of that paper; CEP is
scored on the same windows by the same code.

\paragraph{Probabilistic metrics.} All scores are lower-is-better except coverage.
\begin{itemize}
\item \emph{CRPS} \citep{gneiting2007strictly} measures the accuracy of the whole predictive
distribution; for a point forecast it reduces to the absolute error. We divide it by the average
error of the seasonal-naive forecast on the history, so that values are comparable across series;
a value of $1$ means ``as accurate as seasonal naive was in the past''.
\item \emph{MSIS} \citep{makridakis2020m4} is the Winkler interval score
\citep{winkler1972decision} with the same scaling. It is the width of the interval plus a penalty
each time the outcome falls outside, so it rewards intervals that are narrow and still contain the
outcome.
\item \emph{Winkler score} is the same interval score without the scaling, in the units of the
data. Because it is not comparable across series of different size, we use it only to count wins
within a dataset, as a check that conclusions do not depend on the scaling.
\item \emph{Quantile loss} (normalised mean quantile loss, MQL) is the metric of the GluonTS
library and of the DeepNPTS paper \citep{alexandrov2020gluonts,rangapuram2023deepnpts}, and was
reported in the CSP paper. It is the pinball loss \citep{koenker1978regression} of the forecast
quantiles at levels $0.1,0.2,\dots,0.9$, summed over the horizon, doubled, divided by the sum of
absolute outcomes and averaged over the nine levels. We use the same nine levels and the same
scaling as the CSP paper. Because the levels stop at $0.1$ and $0.9$ it does not see the tails of
the distribution, so we report it as a secondary score.
\item \emph{Coverage} is the share of outcomes inside the $95\%$ interval; the target is $0.95$.
\item \emph{Width} is the average interval width, scaled like MSIS.
\end{itemize}

\paragraph{How methods are compared.} The unit of comparison is the dataset, not the window,
because windows from one dataset are not independent. Scores are first averaged within each
dataset. A method \emph{wins} a dataset when its average is lower. The \emph{ratio} between two
methods is the geometric mean of their per-dataset score ratios: a ratio of $0.86$ means the first
method's score is $14\%$ lower on a typical dataset. Ratios come with a $95\%$ bootstrap interval
($4{,}000$ resamples of datasets), and differences are tested with the paired Wilcoxon signed-rank
test \citep{wilcoxon1945individual}. Mean \emph{ranks} order the methods on each dataset ($1$ is
best) and average the positions.

\paragraph{Development and held-out data.} The corpus was assembled in stages (``tiers''). Design
comparisons were run on the datasets of the first two tiers, the \emph{development set}. The
remaining datasets are divided in two. \emph{Related families} share a source with
a development dataset, for example another frequency of the M4 competition. \emph{Unseen families}
do not.

\paragraph{Fresh data.} As a further check outside the corpus we downloaded $36$ files from $23$
dataset families of fev-bench \citep{shchur2025fevbench}: electricity load and prices, cloud
workloads, retail, epidemiology and macroeconomic panels. We built $1{,}341$ windows from them,
with up to $20$ series per file and up to ten origins per series (three for most). We call this the \emph{fresh data}. DeepNPTS and NPTS were not run on these.

\paragraph{Point-forecast metrics.} A single-number forecast is taken from the samples: the
median for absolute-error metrics and the mean for squared-error metrics. All are lower-is-better.
\emph{MASE} \citep{hyndman2006mase} is the mean absolute error divided by the past error of the
seasonal-naive forecast. \emph{RMSSE} \citep{makridakis2022m5u} is its squared-error counterpart.
\emph{WAPE} is total absolute error divided by total absolute actuals. \emph{Relative MAE} is the
mean absolute error divided by that of a seasonal-naive forecast over the same horizon, so values
below $1$ beat seasonal naive. \emph{Bias} is the mean signed error in MASE units; negative means
the forecasts are too low.

\section{Results}
\label{sec:results}

\paragraph{Summary.} CEP is the best of the six methods on the corpus on both probabilistic
scores (Section~\ref{sec:corpus}). Its advantage over CSP is of similar size on datasets outside
the development set (Section~\ref{sec:heldout}) and on the fresh data (Section~\ref{sec:fresh}). It
is better than Laplace on four datasets in five at a small fraction of the cost
(Sections~\ref{sec:laplace} and~\ref{sec:runtime}), and its point forecasts are more accurate than
those of CSP and Laplace (Section~\ref{sec:point}).

\subsection{The full corpus}
\label{sec:corpus}

Table~\ref{tab:main} gives the scores of the six methods, with every dataset counting equally.
CEP has the best mean rank on CRPS ($1.81$) and on MSIS ($1.77$), and the lowest median CRPS and
MSIS. Its intervals are narrower than CSP's (median width $5.59$ against $6.40$) and still contain
the outcome more often (coverage $0.922$ against $0.901$).

\begin{table}[tbp]
\floatconts
  {tab:main}
  {\caption{Full corpus: $232$ datasets, $24{,}502$ windows, $95\%$ intervals. Lower is better in
  every column except coverage, whose target is $0.95$. Ranks are mean positions among the six
  methods ($1$ is best). CRPS, MSIS and width are medians over datasets; ``MSIS (mean)'' is the
  mean. Rows other than CEP are the stored results of \citet{manokhin2026csp}.}}
  {\resizebox{\linewidth}{!}{{\small\setlength{\tabcolsep}{4pt}
\begin{tabular}{lrrrrrrr}
\toprule
Method & CRPS rank & MSIS rank & CRPS & MSIS & MSIS (mean) & Width & Coverage \\
\midrule
\textbf{\texttt{CEP}} & \textbf{1.81} & \textbf{1.77} & \textbf{0.749} & \textbf{7.98} & \textbf{24.4} & \textbf{5.59} & \textbf{0.922} \\
\texttt{CSP} & 3.06 & 3.22 & 0.841 & 10.23 & 29.4 & 6.40 & 0.901 \\
\texttt{CSP-Nixtla} & 3.36 & 3.51 & 0.842 & 10.17 & 36.3 & 6.10 & 0.872 \\
\texttt{SeasonalNPTS} & 3.99 & 4.06 & 0.900 & 11.38 & 43.3 & 4.79 & 0.763 \\
\texttt{DeepNPTS} & 4.08 & 4.91 & 0.960 & 18.40 & 48.1 & 2.76 & 0.578 \\
\texttt{NPTS} & 4.71 & 3.53 & 1.284 & 10.58 & 41.5 & 7.74 & 0.857 \\
\bottomrule
\end{tabular}}
}}
\end{table}

\begin{table}[tbp]
\floatconts
  {tab:h2h}
  {\caption{CEP against each baseline on the $232$ datasets. ``Wins'' is the share of datasets on
  which CEP has the lower (better) average score; ``Winkler wins'' uses the unscaled Winkler
  score. ``Ratio'' is CEP's score divided by the
  baseline's on a typical dataset (geometric mean), so $0.86$ means $14\%$ lower; the bracket is a
  $95\%$ bootstrap interval. $p$-values are from paired Wilcoxon signed-rank tests.}}
  {\resizebox{\linewidth}{!}{{\small\setlength{\tabcolsep}{4pt}
\begin{tabular}{lrcrcrrr}
\toprule
\texttt{CEP} vs & CRPS wins & CRPS ratio (95\% CI) & MSIS wins & MSIS ratio (95\% CI) & Winkler wins & $p$ (CRPS) & $p$ (MSIS) \\
\midrule
\texttt{CSP} & 78\% & 0.86 (0.83, 0.88) & 84\% & 0.77 (0.74, 0.80) & 83\% & $<10^{-22}$ & $<10^{-24}$ \\
\texttt{CSP-Nixtla} & 81\% & 0.84 (0.80, 0.87) & 87\% & 0.71 (0.67, 0.75) & 86\% & $<10^{-24}$ & $<10^{-29}$ \\
\texttt{DeepNPTS} & 88\% & 0.76 (0.72, 0.80) & 95\% & 0.46 (0.42, 0.50) & 92\% & $<10^{-29}$ & $<10^{-37}$ \\
\texttt{SeasonalNPTS} & 83\% & 0.67 (0.60, 0.73) & 81\% & 0.62 (0.57, 0.67) & 78\% & $<10^{-27}$ & $<10^{-26}$ \\
\texttt{NPTS} & 89\% & 0.57 (0.52, 0.62) & 76\% & 0.66 (0.60, 0.71) & 73\% & $<10^{-33}$ & $<10^{-22}$ \\
\bottomrule
\end{tabular}}
}}
\end{table}

Table~\ref{tab:h2h} compares CEP with each baseline in turn. Against CSP, CEP has the lower
CRPS on $182$ of $232$ datasets ($78\%$), the lower MSIS on $194$ ($84\%$) and the lower unscaled
Winkler score on $193$ ($83\%$). On a typical
dataset its CRPS is $14\%$ lower and its MSIS $23\%$ lower (ratios $0.86$ and $0.77$). Counting
individual windows instead of datasets, CEP has the lower CRPS in $72\%$ of windows and the lower
interval score in $74\%$. Against DeepNPTS, CEP wins $204$ datasets ($88\%$) on CRPS, $220$
($95\%$) on MSIS and $214$ ($92\%$) on the Winkler score. The narrowest margins in the table are
against NPTS on the interval scores: $177$ datasets ($76\%$) on MSIS and $169$ ($73\%$) on the
Winkler score. Scaled and unscaled interval scores give the same picture throughout.

\paragraph{Quantile loss.} Table~\ref{tab:mql} repeats the comparison on the quantile loss. CEP
has the lower quantile loss than CSP on $181$ of $232$ datasets ($78\%$), with a loss $13\%$ lower
on a typical dataset (ratio $0.87$), and than DeepNPTS on $189$ ($81\%$). It wins at least $77\%$
of datasets against every baseline. The quantile loss therefore agrees with CRPS: against CSP the
two scores give $78\%$ of datasets each.

\begin{table}[tbp]
\floatconts
  {tab:mql}
  {\caption{Quantile loss (normalised mean quantile loss over the levels $0.1$ to $0.9$, as in the
  CSP paper; lower is better): CEP against each baseline on the $232$ corpus datasets. ``Median''
  is the median over datasets of the method named in the first column; CEP's median is $0.183$.
  Wins, ratios and $p$-values are read as in Table~\ref{tab:h2h}.}}
  {{\small\setlength{\tabcolsep}{6pt}
\begin{tabular}{lrcrr}
\toprule
Comparison & Wins & Ratio (95\% CI) & Median & $p$ \\
\midrule
\multicolumn{5}{l}{\emph{Corpus: \texttt{CEP} against}}\\
\texttt{CSP} & 78\% & 0.87 (0.84, 0.89) & 0.194 & $<10^{-18}$ \\
\texttt{CSP-Nixtla} & 77\% & 0.88 (0.85, 0.90) & 0.193 & $<10^{-16}$ \\
\texttt{DeepNPTS} & 81\% & 0.87 (0.83, 0.91) & 0.209 & $<10^{-12}$ \\
\texttt{SeasonalNPTS} & 80\% & 0.72 (0.67, 0.77) & 0.267 & $<10^{-21}$ \\
\texttt{NPTS} & 84\% & 0.64 (0.60, 0.69) & 0.297 & $<10^{-23}$ \\
\bottomrule
\end{tabular}}
}
\end{table}

Complete misses are rarer with CEP. The share of windows in which the interval contains none of
the $H$ outcomes is $0.4\%$ for CEP, $1.5\%$ for CSP and $20.4\%$ for DeepNPTS. Coverage is below
$0.85$ on $28$ datasets for CEP and on $34$ for CSP. Figure~\ref{fig:coverage} shows the
distribution of coverage across datasets: CEP removes most of CSP's poorly covered datasets, and
the two methods are similar above $0.85$. Figure~\ref{fig:ratios}(a) shows the per-dataset score
ratios: most datasets lie below $1$, where CEP is better.

\paragraph{How to read Figure~\ref{fig:ratios}.} For each dataset we divide CEP's score by the
other method's score. A ratio of $1$ is a tie; a ratio below $1$ means CEP is the more accurate of
the two on that dataset, and a ratio of $0.5$ means its error is half as large; a ratio above $1$
means the other method is more accurate. The datasets are then sorted from the one most favourable
to CEP to the one least favourable and drawn from left to right, so each curve can only rise. Two
things can be read from a curve. The point at which it crosses the horizontal line at $1$ is the
share of datasets on which CEP wins. How far the curve lies below the line shows how large those
wins are. The solid curve is CRPS and the dashed curve is MSIS. In panel (a), against CSP, the
curves stay below $1$ for $78\%$ of datasets on CRPS and $84\%$ on MSIS, mostly a little below:
CEP wins often and by moderate margins. Panel (b), against Laplace, is discussed in
Section~\ref{sec:laplace}.

\begin{figure}[tbp]
\floatconts
  {fig:coverage}
  {\caption{Coverage of the $95\%$ interval, dataset by dataset. Each curve shows the share of
  datasets (vertical axis) whose coverage is at or below the value on the horizontal axis; the
  dotted line marks the target of $0.95$. A curve that stays low until close to the dotted line
  means few poorly covered datasets. Full corpus, $232$ datasets.}}
  {\includegraphics[width=0.62\linewidth]{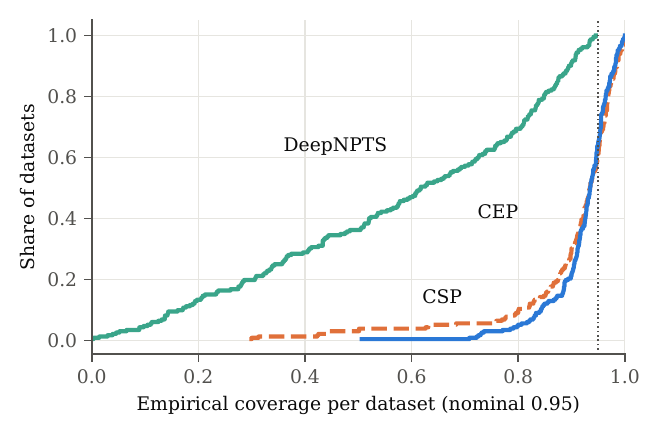}}
\end{figure}

\begin{figure}[tbp]
\floatconts
  {fig:ratios}
  {\caption{CEP's score divided by the other method's score on each dataset, sorted from the
  dataset most favourable to CEP to the least, on a logarithmic scale. Below the horizontal line
  at $1$, CEP is better; the further below, the larger its advantage. The share of the horizontal
  axis over which a curve stays below $1$ is the share of datasets CEP wins. Solid: CRPS; dashed:
  MSIS. (a) CEP against CSP on the full corpus. (b) CEP against Laplace on the
  Laplace subset; the left end is cut off at $0.03$ and contains datasets on which Laplace is
  numerically unstable.}}
  {\includegraphics[width=\linewidth]{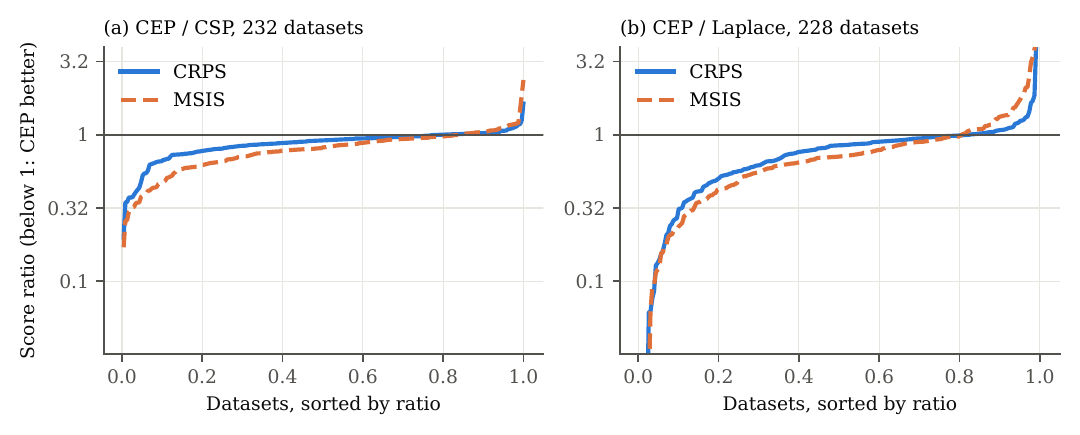}}
\end{figure}

\subsection{Held-out families}
\label{sec:heldout}

A method can look good on the data it was developed on and worse elsewhere. Table~\ref{tab:splits}
therefore splits the corpus into the development set, related families and unseen families. On the
unseen-family datasets CEP beats CSP on $76\%$ by CRPS, on $83\%$ by MSIS, on $81\%$ by Winkler
score and on $76\%$ by quantile loss. Its CRPS is $15\%$ lower and its MSIS $22\%$ lower on a typical dataset (ratios $0.85$ and
$0.78$, both $p<10^{-13}$). The related families give similar ratios ($0.89$ and $0.79$). The
development set gives larger gains ($0.80$ and $0.66$). The coverage gain is concentrated where CSP was weak: on unseen families CSP already
covers $0.915$ and CEP reaches $0.922$.

\begin{table}[tbp]
\floatconts
  {tab:splits}
  {\caption{CEP against CSP by relation to the development data. Development: the datasets on
  which design comparisons were run. Related: other datasets from the same sources. Unseen: all
  remaining datasets. Wins are the share of datasets in the split on which CEP has the lower
  score; ratios are read as in Table~\ref{tab:h2h}.}}
  {\resizebox{\linewidth}{!}{{\small\setlength{\tabcolsep}{4pt}
\begin{tabular}{lrrrcrcr}
\toprule
Split & \multicolumn{2}{c}{Coverage} & CRPS wins & CRPS ratio (95\% CI) & MSIS wins & MSIS ratio (95\% CI) & Winkler wins \\
\cmidrule(lr){2-3}
 & CEP & CSP & & & & & \\
\midrule
Development & 0.941 & 0.871 & 86\% & 0.80 (0.73, 0.88) & 86\% & 0.66 (0.55, 0.78) & 86\% \\
Related families & 0.917 & 0.874 & 83\% & 0.89 (0.86, 0.92) & 83\% & 0.79 (0.73, 0.85) & 86\% \\
Unseen families & 0.922 & 0.915 & 76\% & 0.85 (0.81, 0.89) & 83\% & 0.78 (0.74, 0.82) & 81\% \\
\bottomrule
\end{tabular}}
}}
\end{table}

\subsection{Fresh data}
\label{sec:fresh}

On the $36$ fresh files (Table~\ref{tab:fresh}) CEP has the lower CRPS on $28$ ($78\%$) and the
lower MSIS on $29$ ($81\%$), with $p=0.002$ and $p<10^{-4}$; it also has the lower Winkler score
on $29$ and the lower quantile loss on $26$ ($72\%$, $p=0.006$). On a typical file its CRPS is $16\%$
lower and its MSIS $25\%$ lower (ratios $0.84$, interval $0.74$--$0.93$, and $0.75$, interval
$0.66$--$0.85$). Coverage is $0.904$ against $0.877$.

\begin{table}[tbp]
\floatconts
  {tab:fresh}
  {\caption{Fresh data: $36$ files from $23$ families, $1{,}341$ windows. CRPS, MSIS and width are
  medians over files; lower is better. Wins are the share of files on which CEP has the lower score than CSP; $p$-values are for CEP
  against CSP.}}
  {\resizebox{\linewidth}{!}{{\small\setlength{\tabcolsep}{4pt}
\begin{tabular}{lrrrrrrrrr}
\toprule
Method & Coverage & CRPS & MSIS & Width & CRPS wins & MSIS wins & Winkler wins & $p$ (CRPS) & $p$ (MSIS) \\
\midrule
\texttt{CSP} & 0.877 & 1.016 & 15.52 & 7.29 & -- & -- & -- & -- & -- \\
\texttt{CEP} & 0.904 & 0.886 & 13.27 & 7.33 & 78\% & 81\% & 81\% & $0.002$ & $<10^{-4}$ \\
\bottomrule
\end{tabular}}
}}
\end{table}

\subsection{Comparison with Laplace}
\label{sec:laplace}

\paragraph{The published claims and what they rest on.} The author of Laplace states that
``Laplace beats (almost) everything'' and that it gives ground to CSP only ``narrowly'', on strongly
seasonal data, attributing that loss to CSP being given the seasonal period
\citep{cotton2026skaters}. The supporting comparison reports Laplace ``beating'' CSP on $82$ to $99\%$ of
series in three collections and losing on the fourth \citep{cotton2026laplacevscsp}.
Table~\ref{tab:protocols} sets that comparison beside ours. It is narrow in three ways.
\begin{itemize}
\item \emph{One step ahead only.} Every forecast is for the next observation. Multi-step
forecasting, the usual practical task, is not tested.
\item \emph{A limited and unrepresentative choice of data.} Three of the four collections are
economic change series from a single source (FRED); the fourth is M4-Hourly. The series were chosen
by Laplace's author. A handful of collections picked by the author of the method under test, almost all of them from
one source, is not an acceptable basis for a claim of general superiority. It falls well short of
professional benchmarking practice and of the current standard in forecasting, which is
large-scale evaluation on many datasets from many domains and frequencies, fixed in advance and
independent of the method being promoted: the M competitions
\citep{makridakis2020m4,makridakis2022m5u}, the Monash archive \citep{godahewa2021monash},
GIFT-Eval \citep{aksu2024gifteval}, TFB \citep{qiu2024tfb}, ProbTS \citep{zhang2024probts} and
fev-bench, the benchmark maintained by Amazon \citep{shchur2025fevbench}. Professional
practice, as those benchmarks embody it, has four requirements, and the comparison meets none of
them. The data are assembled by someone other than the author of the method being tested, and
before it is tested. They cover many domains, frequencies and series lengths, so that no single
kind of series decides the result. Forecasts are made at the horizons at which forecasts are
actually used, not one step ahead only. And every standard measure is reported, including those on
which the method does badly. A comparison that fails all four is a demonstration prepared by the
method's author, not a benchmark.
\item \emph{Two metrics.} Only CRPS and log-likelihood are reported. Interval width, interval score
and, for CSP, coverage are not reported. This departs from accepted practice: probabilistic
forecasts are evaluated by maximising sharpness, that is the narrowness of the predictive
intervals, subject to calibration, so width and coverage are reported together
\citep{gneiting2007calibration,gneiting2007strictly,gneiting2014probabilistic}. On our data these are unfavourable to Laplace: its
intervals are $44\%$ wider than CSP's (median scaled width $9.40$ against $6.54$).
\end{itemize}
A result obtained this way does not support a claim of general superiority. Stated directly: the
claim that Laplace ``beats (almost) everything'' is false, and the evidence offered for it is
cherry-picked twice over, in the data and in the metrics. The data are a handful of collections
chosen by the method's own author, almost all from one source. The metrics are CRPS and
log-likelihood only. CRPS is an accepted proper scoring rule \citep{gneiting2007strictly}. The
log-likelihood is also a proper score, but it is not what forecasting is benchmarked on: none of
the benchmarks cited above uses it, it can only be computed for methods that output a density, so
that sample-based forecasters such as CSP must first be smoothed into one, and it is dominated by
the density assigned to a few extreme outcomes. What the comparison leaves out is what the field
has agreed on for decades. Calibration and sharpness, that is the coverage and the width of
prediction intervals, are the accepted pair of criteria for a probabilistic forecast
\citep{gneiting2007calibration,gneiting2014probabilistic}; interval widths are not reported. The
Winkler interval score, which combines the two in a single number, has been in use since
\citet{winkler1972decision} and is the interval metric of the M4 competition
\citep{makridakis2020m4}; it is not reported either. A study that selects its own data, omits
interval width and omits the standard interval score cannot establish that one forecaster is
better than another, and on these omitted measures Laplace is the worse of the two.

\begin{table}[tbp]
\floatconts
  {tab:protocols}
  {\caption{The comparison of Laplace with CSP published by Laplace's author
  \citep{cotton2026laplacevscsp} and the comparison in this paper.}}
  {{\small
\begin{tabular}{p{0.19\linewidth}p{0.36\linewidth}p{0.36\linewidth}}
\toprule
 & Comparison published by Laplace's author & This paper \\
\midrule
Forecast horizon & One step ahead only & Multi-step, $1$ to $168$ steps \\
Data & $1{,}409$ series in four collections: daily, weekly and monthly FRED economic change series, and M4-Hourly & $228$ datasets: energy, transport, retail, web traffic, finance, economics, health, weather, cloud workloads, M-competitions \\
Choice of data & By Laplace's author & The CSP corpus, fixed beforehand \\
Probabilistic metrics & CRPS and log-likelihood & CRPS, MSIS, Winkler score, coverage, interval width \\
Interval width, interval score & Not reported & Reported \\
Point accuracy & Not reported & MASE, RMSSE, bias \\
Runtime & Not reported for this comparison & Reported \\
History & Each method's own & Identical for all methods \\
Result & Laplace ``wins'' $82$--$99\%$ of FRED series by CRPS and loses M4-Hourly & CSP wins $60\%$ of datasets by CRPS and $64\%$ by interval score \\
\bottomrule
\end{tabular}}
}
\end{table}

\paragraph{The claims tested on the CSP benchmark.} We ran Laplace on the CSP benchmark, with
multi-step horizons and identical history for every method: $2{,}534$ windows from $228$ of its
datasets, which we call the \emph{Laplace subset}. The claims of Laplace's author do not hold on large-scale
studies such as this one. Plain CSP has the lower CRPS than Laplace on $137$ of $228$ datasets
($60\%$) and the lower interval score on $147$ ($64\%$), by both MSIS and the Winkler score
($p<10^{-5}$ and $p<10^{-4}$). Its point forecasts are also more accurate: CSP has the lower MASE
on $128$ datasets ($56\%$) and the lower RMSSE on $120$ ($53\%$). CSP's advantage is at
multi-step horizons, which a one-step test cannot see.

\paragraph{Cost.} Every forecast is timed from a cold start: for Laplace this includes reading its
warm-up observations. CSP needs $1.9$ seconds for the $2{,}534$ forecasts and Laplace $14{,}664$
seconds, about $7{,}700$ times longer. The gap widens with the horizon. Per forecast, Laplace is
about $2{,}500$ times slower than CSP at horizons up to $12$ and about $24{,}000$ times slower
beyond $96$, where one forecast takes $107$ seconds
(Table~\ref{tab:speed}). Its cost per observation grows roughly with the square of the horizon. At
the longest horizons Laplace is impractical to run: this is why only $2{,}534$ of the benchmark's
$24{,}502$ windows could be included and why the dataset with
$H=336$ was left out, where by our cost fit a single forecast would take about six minutes. The
same holds for the standard forecasting competitions. Timed on samples of their series, Laplace
would need about $52$ processor-hours for the $100{,}000$ series of M4 \citep{makridakis2020m4}
and about $105$ processor-hours for the $42{,}840$ series of the M5 uncertainty track
\citep{makridakis2022m5u}, at $8.8$ seconds per M5 series; CSP forecasts all of M4 in $66$ seconds
and all of M5 in $130$ seconds.

\paragraph{CEP against Laplace.} Table~\ref{tab:laplace} compares Laplace, CSP and CEP when all three are given the same
history. CEP has the lower CRPS on $187$ of $228$ datasets ($82\%$), the lower MSIS on $183$
($80\%$) and the lower Winkler score on $186$ ($82\%$). On a typical dataset its CRPS is $43\%$ lower than Laplace's and its MSIS $48\%$ lower
(ratios $0.57$ and $0.52$). Its intervals are $41\%$ narrower at the median ($5.57$ against
$9.40$). 

\begin{table}[tbp]
\floatconts
  {tab:laplace}
  {\caption{Comparison with Laplace on $228$ corpus datasets, with the same
  history given to all three methods. CRPS, MSIS and width are medians over datasets; lower is
  better. Wins are the share of datasets on which the method has the lower score than Laplace;
  $p$-values are for each method against Laplace.}}
  {\resizebox{\linewidth}{!}{{\small\setlength{\tabcolsep}{4pt}
\begin{tabular}{lrrrrrrrr}
\toprule
Method & CRPS & MSIS & Width & CRPS wins & MSIS wins & Winkler wins & $p$ (CRPS) & $p$ (MSIS) \\
\midrule
\texttt{CEP} & 0.756 & 7.52 & 5.57 & 82\% & 80\% & 82\% & $<10^{-20}$ & $<10^{-15}$ \\
\texttt{CSP} & 0.824 & 10.12 & 6.54 & 60\% & 64\% & 64\% & $<10^{-5}$ & $<10^{-4}$ \\
\texttt{Laplace} & 0.960 & 11.68 & 9.40 & -- & -- & -- & -- & -- \\
\bottomrule
\end{tabular}}
}}
\end{table}

\paragraph{Laplace among all methods.} Table~\ref{tab:laplaceall} reports Laplace in the same way
as the other methods of the benchmark: all seven methods are scored on the $228$ datasets on which
Laplace was run, with ranks computed among the seven. {\itshape \textbf{The result is stark. Laplace is near the
bottom of the table: it ranks sixth of seven on CRPS and sixth of seven on MSIS.} On CRPS it ranks
below every method except plain NPTS, including SeasonalNPTS, a kernel sampler with no model at
all, and DeepNPTS, the method that CSP had already beaten. On MSIS it ranks below every method
except DeepNPTS, and so below both NPTS and SeasonalNPTS. CEP, CSP and CSP-Nixtla are each ahead of
it on both scores. \textbf{Its intervals are also by far the widest of the seven methods:} a median width
of $9.40$, against $7.15$ for the next widest (NPTS), $6.54$ for CSP and $5.57$ for CEP, that is
$31\%$, $44\%$ and $69\%$ wider. Laplace buys its coverage with width and still loses on the
interval score. \textbf{A
forecaster said to beat almost everything finishes behind elementary non-parametric samplers on a
benchmark of $228$ datasets.}}

\begin{table}[tbp]
\floatconts
  {tab:laplaceall}
  {\caption{All methods on the $228$ datasets ($2{,}534$ windows) on which Laplace was run. Columns
  as in Table~\ref{tab:main}, without coverage; ranks are mean positions among the seven methods. ``Wins'' is the
  share of datasets on which the method has the lower score than Laplace. Laplace, CSP and CEP
  receive the same truncated history; the rows of CSP-Nixtla, DeepNPTS, NPTS and SeasonalNPTS are
  the stored results of \citet{manokhin2026csp} on the same windows, with the full history.}}
  {\resizebox{\linewidth}{!}{{\small\setlength{\tabcolsep}{4pt}
\begin{tabular}{lrrrrrrr}
\toprule
Method & CRPS rank & MSIS rank & CRPS & MSIS & Width & CRPS wins & MSIS wins \\
\midrule
\textbf{\texttt{CEP}} & \textbf{2.23} & \textbf{2.38} & \textbf{0.756} & \textbf{7.52} & \textbf{5.57} & \textbf{82\%} & \textbf{80\%} \\
\texttt{CSP-Nixtla} & 3.75 & 3.94 & 0.819 & 9.45 & 5.93 & 58\% & 61\% \\
\texttt{CSP} & 3.81 & 3.73 & 0.824 & 10.12 & 6.54 & 60\% & 64\% \\
\texttt{DeepNPTS} & 4.36 & 5.22 & 0.973 & 18.28 & 2.50 & 50\% & 38\% \\
\texttt{SeasonalNPTS} & 4.45 & 4.19 & 0.920 & 10.73 & 4.61 & 51\% & 59\% \\
\texttt{Laplace} & 4.50 & 4.67 & 0.960 & 11.68 & 9.40 & -- & -- \\
\texttt{NPTS} & 4.91 & 3.87 & 1.200 & 10.12 & 7.15 & 48\% & 64\% \\
\bottomrule
\end{tabular}}}}
\end{table}

The picture varies by dataset (Figure~\ref{fig:ratios}b, read as explained in
Section~\ref{sec:corpus}). The curves stay below $1$, where CEP is the more accurate, for $82\%$
of datasets on CRPS and $80\%$ on MSIS, and they lie much further below $1$ than in panel (a): the
wins over Laplace are larger than the wins over CSP. The part of each curve above $1$, at the
right, is the minority of datasets on which Laplace is the more accurate. CEP's CRPS is less than half of Laplace's on $45$
datasets ($20\%$) and less than a tenth of it on $9$. On these Laplace fails in one of two ways: on Bitcoin and CIF 2016 its forecast distribution becomes
numerically unstable and produces enormous values, and on electricity load (Australian, Taylor and
Zurich) and M4-Hourly it abandons the cycle and forecasts negative values. These cases push Laplace's \emph{mean} MSIS above $10^{14}$, which is why
Table~\ref{tab:laplace} reports medians.

\paragraph{Pathological forecasts.} Win rates do not show what a forecast looks like.
Figure~\ref{fig:pathological} shows five of Laplace's forecasts from the benchmark, with CSP's
interval from the same history. On three electricity series (Australian half-hourly demand, ERCOT
West and the Taylor load series) Laplace's forecast leaves the daily cycle. On the Australian and
Taylor series its median falls roughly linearly and becomes negative within a day; on ERCOT West
the lower end of its $95\%$ interval is negative at $19$ of the $24$ steps. On
the Taylor series, where load lies between $21{,}000$ and $37{,}000$, the lower end of its $95\%$
interval reaches $-109{,}000$. On an M4-Hourly series that lies between $294$ and $800$ its interval
runs from $-4{,}242$ to $1{,}763$, and its median forecast has a mean absolute error of $874$ against
$26$ for CSP. On the Bitcoin hash rate, of order $10^{20}$, its interval reaches $10^{35}$. The four seasonal cases share one pattern: at
multi-step horizons Laplace does not carry the cycle forward but extends the last movement of the
series and widens the interval with the horizon, as a random walk would.

Two counts show that these are not isolated cases. First, in $2{,}460$ of the $2{,}534$ windows
every outcome is non-negative; Laplace's median forecast goes below zero in $187$ of them ($7.6\%$),
against $12$ for CSP and $2$ for CEP. Second, Laplace's point forecast is less accurate than the
seasonal-naive forecast over the same horizon on $122$ of $227$ datasets ($54\%$), against $76$
($33\%$) for CSP and $57$ ($25\%$) for CEP.

\begin{figure}[tbp]
\floatconts
  {fig:pathological}
  {\caption{Examples of Laplace forecasts on the benchmark (\texttt{skaters} 0.16.0, default
  settings). Each panel shows the end of the history, the outcome,
  Laplace's median forecast and $95\%$ interval, and CSP's $95\%$ interval from the same history.
  The horizontal axis counts steps from the forecast origin. The bottom panel uses a logarithmic
  axis.}}
  {\includegraphics[width=\linewidth]{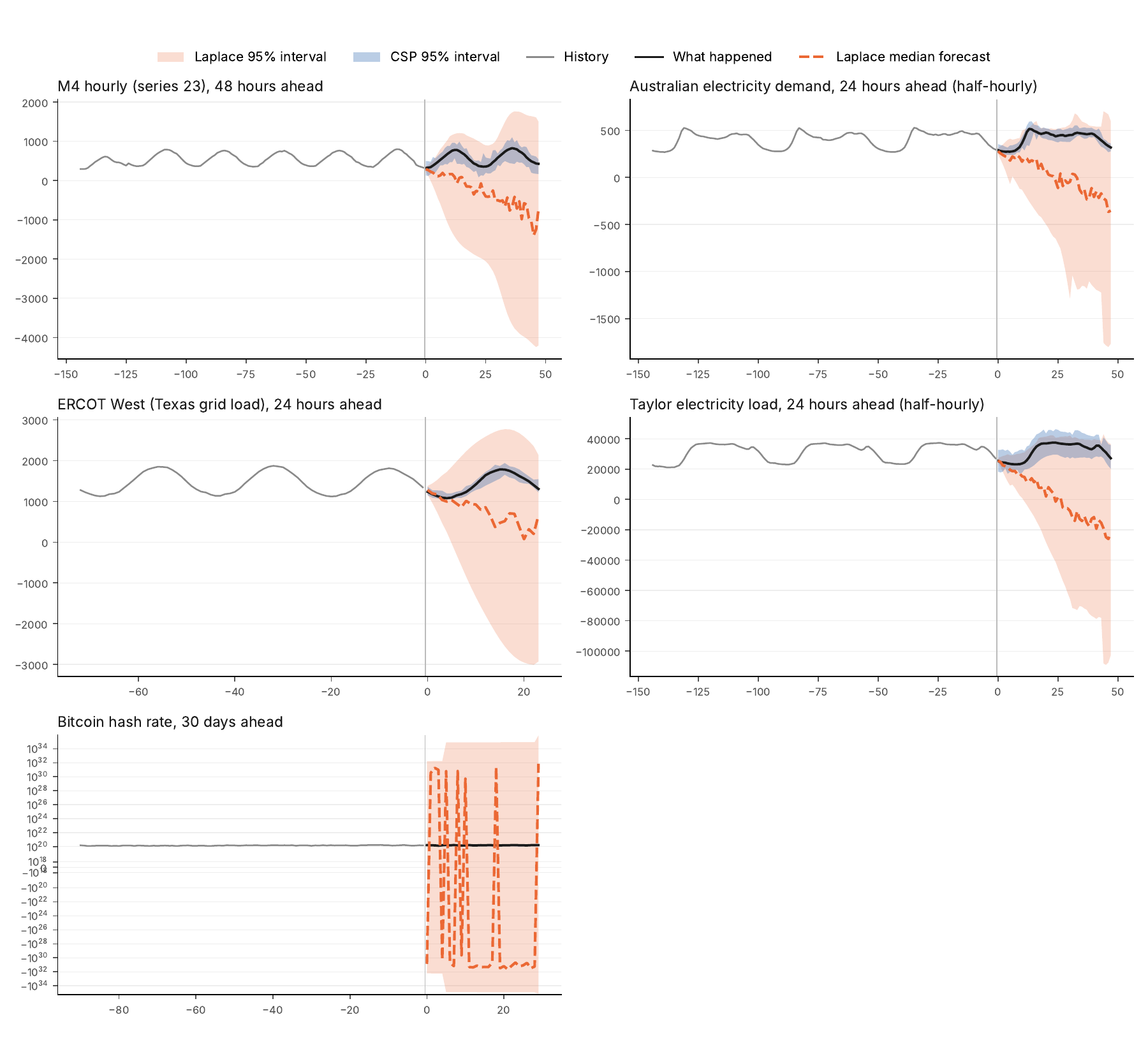}}
\end{figure}

On the fresh data, again with the same history for all methods (Table~\ref{tab:freshlaplace}), CEP
has the lower CRPS on $29$ of $36$ files ($81\%$), the lower MSIS on $25$ ($69\%$) and the lower
Winkler score on $26$ ($72\%$). The median
MSIS is $9.60$ for CEP against $19.26$ for Laplace.

\begin{table}[tbp]
\floatconts
  {tab:freshlaplace}
  {\caption{Comparison with Laplace on the fresh data: $407$ windows from $36$ files, same history
  for all three methods. CRPS, MSIS and width are medians over files; wins are the share of files
  on which the method has the lower score than Laplace.}}
  {{\small\setlength{\tabcolsep}{4pt}
\begin{tabular}{lrrrrrr}
\toprule
Method & CRPS & MSIS & Width & CRPS wins & MSIS wins & Winkler wins \\
\midrule
\texttt{CEP} & 0.948 & 9.60 & 7.36 & 81\% & 69\% & 72\% \\
\texttt{CSP} & 0.995 & 10.54 & 7.78 & 61\% & 53\% & 53\% \\
\texttt{Laplace} & 1.046 & 19.26 & 12.58 & -- & -- & -- \\
\bottomrule
\end{tabular}}
}
\end{table}

\paragraph{Summary of the evidence on Laplace.} Every measure reported in this section points the
same way. Laplace is less accurate than CSP on most datasets by CRPS, by interval score and by
point error. Its intervals are the widest of all the methods compared. It ranks sixth of seven
methods. It produces forecasts that are impossible for the data, and it becomes numerically
unstable on some series. It costs thousands of times more to run than the baseline it loses to.
None of these findings depends on a single dataset or a single metric, and none of them is visible
in the comparison published by its author, because that comparison does not measure them.

\subsection{Point-forecast accuracy}
\label{sec:point}

A probabilistic forecaster should also give a good single-number forecast.
Table~\ref{tab:point} reports point metrics. On the full corpus, CEP lowers the median MASE from
$1.10$ (CSP) to $1.01$ and the median RMSSE from $0.94$ to $0.87$. It wins $181$ of $232$ datasets
($78\%$) on MASE and $177$ ($76\%$) on RMSSE, with errors about $11\%$ and $12\%$ lower on a
typical dataset (ratios $0.89$ and $0.88$, both $p<10^{-13}$). Its absolute error is $25\%$ below
that of a seasonal-naive forecast over the same horizons.

On the Laplace subset, CEP has the lower MASE on $162$ of $228$ datasets ($71\%$) and the lower
RMSSE on $160$ ($70\%$). \textbf{\itshape Laplace's point forecast is on average worse than seasonal naive on these
windows (relative MAE $1.22$) and tends to be too low (bias $-0.24$, against $-0.02$ for CEP).}
Absence of bias is a key requirement of any forecaster, because a bias does not average out: it is
the same error made again and again in the same direction. A forecast that is systematically too
low makes a retailer or manufacturer order and stock too little, which means stock-outs and lost
sales; it makes an energy supplier buy or bid too little for the grid, which means buying the
shortfall at imbalance prices; and no amount of interval width repairs it, because the whole
distribution is centred in the wrong place. {\itshape Laplace fails this requirement: its bias is
$-0.24$, against $-0.02$ for CSP and CEP on the same windows.}

\begin{table}[tbp]
\floatconts
  {tab:point}
  {\caption{Point-forecast accuracy on the four test sets. Lower is better for MASE, RMSSE, WAPE and
  relative MAE; bias should be near zero. MASE, RMSSE, WAPE and bias are medians over datasets;
  relative MAE (``rel.\ MAE'') is a geometric mean, with values below $1$ beating seasonal naive.
  Within each block methods are listed from most to least accurate. Wins are the share of datasets (or files) on which the method is more accurate than the last
  method in its block.}}
  {\resizebox{\linewidth}{!}{{\small\setlength{\tabcolsep}{4pt}
\begin{tabular}{llrrrrrrr}
\toprule
Test set & Method & MASE & RMSSE & WAPE & rel.\ MAE & Bias & MASE wins & RMSSE wins \\
\midrule
Corpus, full history ($n=232$) & \texttt{CEP} & 1.012 & 0.867 & 0.170 & 0.748 & -0.091 & 78\% & 76\% \\
 & \texttt{CSP} & 1.101 & 0.945 & 0.183 & 0.875 & -0.060 & -- & -- \\
\midrule
Corpus, Laplace subset ($n=228$) & \texttt{CEP} & 1.021 & 0.867 & 0.150 & 0.772 & -0.022 & 71\% & 70\% \\
 & \texttt{CSP} & 1.128 & 0.960 & 0.180 & 0.870 & -0.018 & 56\% & 53\% \\
 & \texttt{Laplace} & 1.213 & 1.052 & 0.260 & 1.222 & -0.238 & -- & -- \\
\midrule
Fresh, full history ($n=36$) & \texttt{CEP} & 1.194 & 0.990 & 0.167 & 0.871 & -0.071 & 75\% & 69\% \\
 & \texttt{CSP} & 1.434 & 1.188 & 0.163 & 0.984 & -0.093 & -- & -- \\
\midrule
Fresh, Laplace subset ($n=36$) & \texttt{CEP} & 1.187 & 1.010 & 0.140 & 0.886 & -0.052 & 78\% & 81\% \\
 & \texttt{CSP} & 1.363 & 1.121 & 0.159 & 0.987 & -0.209 & 56\% & 56\% \\
 & \texttt{Laplace} & 1.469 & 1.173 & 0.248 & 1.581 & -0.508 & -- & -- \\
\bottomrule
\end{tabular}}}}
\end{table}

\subsection{Runtime}
\label{sec:runtime}

On the full corpus CEP needs $1{,}778$ seconds of computing time on one processor core for all
$24{,}502$ forecasts, a median of $39$ milliseconds per forecast; CSP needs $40$ seconds in total.
Table~\ref{tab:speed} gives the median time per forecast on the Laplace subset, with all three
methods timed in the same run. For Laplace the time is that of reading the warm-up observations
one by one and then producing the forecast. Depending on the horizon, CEP is between $77$ and
about $1{,}300$ times faster than Laplace, and roughly $20$ to $30$ times slower than CSP.

\begin{table}[tbp]
\floatconts
  {tab:speed}
  {\caption{Median time per forecast in milliseconds on the Laplace subset, by forecast horizon,
  with totals in seconds in the last row. Every forecast is timed from a cold start. The last two columns are Laplace's time divided by CEP's and by CSP's.}}
  {\resizebox{\linewidth}{!}{{\small
\begin{tabular}{lrrrrr}
\toprule
Horizon $H$ & \texttt{CSP} (ms) & \texttt{CEP} (ms) & \texttt{Laplace} (ms) & Laplace / CEP & Laplace / CSP \\
\midrule
1--12 & 0.3 & 9.8 & 750 & 77$\times$ & 2{,}500$\times$ \\
13--24 & 0.9 & 21.4 & 5{,}030 & 235$\times$ & 5{,}600$\times$ \\
25--48 & 1.2 & 38.8 & 11{,}755 & 303$\times$ & 9{,}800$\times$ \\
49--96 & 2.4 & 59.0 & 47{,}304 & 802$\times$ & 19{,}700$\times$ \\
$>$96 & 4.5 & 82.4 & 107{,}024 & 1299$\times$ & 23{,}800$\times$ \\
\midrule
Total, 2{,}534 windows (s) & 1.9 & 54.1 & 14{,}664 & 271$\times$ & 7{,}700$\times$ \\
\bottomrule
\end{tabular}}
}}
\end{table}

\section{Conclusion}

CSP is deliberately minimal: it has a single forecast of the level of a series, the seasonal-naive
forecast. CEP replaces it by a pool of simple forecasters whose past errors, adapted to the series,
form the predictive distribution. The result is the strongest method on the benchmark. Among the
six methods compared on $232$ datasets CEP has the best mean rank on both CRPS and the interval
score. It improves on CSP on $78\%$ of datasets by CRPS and on $84\%$ by interval score, with
intervals that are narrower and cover more often. It beats DeepNPTS, a trained neural forecaster,
on $88\%$ of datasets by CRPS and on $95\%$ by interval score. The gains hold on datasets
outside the development tiers and on data from outside the benchmark, and its point forecasts
are more accurate as well. CEP achieves this with no training, no learned parameters and a median
of $39$ milliseconds per forecast on a single processor core. It shows how far the training-free,
conformal approach can be taken, and it sets the standard that a trained probabilistic forecaster
should be required to beat.

The evidence on Laplace is unambiguous. The claim that it ``beats (almost) everything'' is false.
On $228$ datasets it loses to CSP, a baseline of a few lines of numpy code, on $60\%$ by CRPS and
on $64\%$ by interval score, and it loses to CEP on four datasets in five. Placed among all the
methods of the benchmark it ranks sixth of seven, behind elementary kernel samplers. Its intervals
are the widest of any method, $44\%$ wider than CSP's and $69\%$ wider than CEP's. Its point
forecast is worse than the seasonal-naive forecast on more than half of the datasets and is biased
low ($-0.24$ in scaled units, against $-0.02$ for CSP and CEP). It forecasts negative values for series that cannot be negative,
abandons seasonal cycles and, on some series, returns intervals fifteen orders of magnitude wider
than the data. From a cold start it is about $7{,}700$ times slower than CSP and would need days of processor time
for a single pass over M4 or M5. Its seasonal anchor was taken from CSP without credit. The
published case for it rests on a handful of series chosen by its own author, scored one step ahead,
with interval width and the interval score left out. On this evidence Laplace is not a competitive
forecaster, and it should not be presented as one.

\bibliography{refs}

\end{document}